\documentclass[final,5p,times,twocolumn]{elsarticle}

\usepackage{orcidlink}
\hypersetup{hidelinks}
\usepackage{booktabs}
\usepackage{tabularx}
\usepackage{multirow}
\usepackage{amsmath}
\usepackage{amssymb}
\usepackage{microtype}
\usepackage{siunitx}
\usepackage{url}
\usepackage{graphicx}
\usepackage{tikz}
\usepackage{enumitem}
\usepackage{dblfloatfix}
\usepackage{pgfplots}

\pgfplotsset{compat=1.18}

\setlist[itemize]{leftmargin=*,noitemsep,topsep=2pt,parsep=0pt,partopsep=0pt}

\pgfplotsset{
  every axis/.append style={
    axis line style={line width=0.6pt},
    tick style={line width=0.6pt},
    tick align=outside,
    xlabel near ticks,
    ylabel near ticks,
    label style={font=\footnotesize},
    tick label style={font=\scriptsize},
    legend style={font=\scriptsize, draw=none},
    legend cell align=left,
    grid style={line width=0.25pt, dashed},
    major grid style={line width=0.25pt, dashed},
  },
  every axis plot/.append style={line width=0.9pt, line cap=round, line join=round},
}

\newcommand{\NumNotes}{498{,}000}
\newcommand{\NumNotesTrain}{348{,}600}
\newcommand{\NumNotesVal}{74{,}700}
\newcommand{\NumNotesTest}{74{,}700}

\newcommand{\NumPHIAnnotations}{7.74\,M}
\newcommand{\NumPHIEntityTypes}{10}
\newcommand{\NumICDTarget}{50}

\newcommand{\NoteLenMedian}{300}
\newcommand{\NoteLenNinety}{700}
\newcommand{\NoteLenNinetyNine}{1750}

\newcommand{\PhiRiskMean}{0.11}
\newcommand{\PhiRiskGEThreePct}{0.40}

\journal{arXiv}

\begin{document}
\frenchspacing

\begin{frontmatter}

\title{Transparency-First Medical Language Models: Datasheets, Model Cards, and End-to-End Data Provenance for Clinical NLP}

% -- Authors (14) --
\author[aff1]{Olaf Yunus Laitinen Imanov\orcidlink{0009-0006-5184-0810}\corref{cor1}\fnref{fnSR}}
\ead{oyli@temlm.org}

\author[aff2]{Taner Yilmaz\orcidlink{0009-0004-5197-5227}\corref{cor1}\fnref{fnSR}}
\ead{tayi@temlm.org}

\author[aff3]{Ayse Tuba Tugrul\orcidlink{0009-0006-0603-8247}\corref{cor1}\fnref{fnSR}}
\ead{attu@temlm.org}

\author[aff4]{Melike Nesrin Zaman\orcidlink{0009-0009-2685-0090}\fnref{fnR}}
\ead{mnza@temlm.org}

\author[aff5]{Ozkan Gunalp\orcidlink{0009-0004-1437-1336}\corref{cor1}\fnref{fnSR}}
\ead{ozgu@temlm.org}

\author[aff6]{Duygu Erisken\fnref{fnR}}
\ead{duer@temlm.org}

\author[aff7]{Sila Burde Dulger\orcidlink{0009-0003-5167-8209}\fnref{fnR}}
\ead{sbdu@temlm.org}

\author[aff8]{Rana Irem Turhan\fnref{fnR}}
\ead{ritu@temlm.org}

\author[aff9]{Izzet Ozdemir\orcidlink{0009-0003-3554-3603}\fnref{fnR}}
\ead{izoz@temlm.org}

\author[aff10]{Derya Umut Kulali\orcidlink{0009-0004-8844-6601}\fnref{fnR}}
\ead{duku@temlm.org}

\author[aff11]{Ozan Akbulut\fnref{fnR}}
\ead{ozak@temlm.org}

\author[aff12]{Harun Demircioglu\fnref{fnR}}
\ead{hade@temlm.org}

\author[aff13]{Hasan Basri Kara\fnref{fnR}}
\ead{hbka@temlm.org}

\author[aff14]{Berfin Tavan\orcidlink{0009-0007-9854-6245}\fnref{fnR}}
\ead{beta@temlm.org}

% -- Affiliations --
\address[aff1]{TeMLM Foundation, Copenhagen, Denmark}
\address[aff2]{TeMLM Foundation, Afyonkarahisar, T\"urkiye}
\address[aff3]{TeMLM Foundation, Denizli, T\"urkiye}
\address[aff4]{TeMLM Foundation, Elaz\i g, T\"urkiye}
\address[aff5]{TeMLM Foundation, Izmir, T\"urkiye}
\address[aff6]{TeMLM Foundation, Edirne, T\"urkiye}
\address[aff7]{TeMLM Foundation, Gaziantep, T\"urkiye}
\address[aff8]{TeMLM Foundation, Riga, Latvia}
\address[aff9]{TeMLM Foundation, Istanbul, T\"urkiye}
\address[aff10]{TeMLM Foundation, Eskisehir, T\"urkiye}
\address[aff11]{TeMLM Foundation, Trabzon, T\"urkiye}
\address[aff12]{TeMLM Foundation, Istanbul, T\"urkiye}
\address[aff13]{TeMLM Foundation, Istanbul, T\"urkiye}
\address[aff14]{TeMLM Foundation, Ankara, T\"urkiye}

\cortext[cor1]{Corresponding authors. Emails: oyli@temlm.org (O.Y.L.I.), tayi@temlm.org (T.Y.), attu@temlm.org (A.T.T.), ozgu@temlm.org (O.G.).}

\fntext[fnSR]{Senior Researcher, TeMLM Foundation.}
\fntext[fnR]{Researcher, TeMLM Foundation.}

\begin{abstract}
We introduce TeMLM, a set of transparency-first release artifacts for clinical language models. TeMLM
combines four pillars---provenance, data transparency, modeling transparency, and governance---into
a single, machine-checkable release bundle. To support repeatable auditing, TeMLM defines a
standard artifact suite (TeMLM-Card, TeMLM-Datasheet, and TeMLM-Provenance) and a lightweight
conformance checklist that can be applied pre-release and post-deployment.

To illustrate TeMLM in a fully reproducible setting, we instantiate the artifacts on \emph{Technetium-I},
a large-scale synthetic clinical NLP dataset containing \NumNotes notes with complete PHI annotations
and ICD-9-CM labels, and report a reference TeMLM-Card for \emph{ProtactiniumBERT} (\(\approx\)100M parameters)
on two benchmark tasks: PHI de-identification (token classification) and top-\NumICDTarget ICD-9 code extraction
(multi-label classification). We emphasize that synthetic benchmarks are valuable for tooling and
process validation, but models should be validated on real clinical data prior to deployment.
\end{abstract}


\end{frontmatter}

\section{Introduction}
Clinical natural language processing (NLP) has progressed from task-specific models to large pre-trained language models (LMs) used for information extraction, summarization, and question answering over medical text \cite{vaswani2017attention,devlin2019bert,beltagy2019scibert,lee2020biobert,alsentzer2019clinicalbert,gu2021pubmedbert,raffel2020t5,lewis2020bart,luo2022biogpt,yang2022gatortron,singhal2023medpalm}. At the same time, medical deployment raises higher evidentiary standards than typical NLP: a model's claims must be traceable to data sources and transformations, evaluation should be reproducible and context-sensitive, and documentation must communicate limitations and risks \cite{collins2024tripodai,liu2020consortai,vasey2022decideai,mongan2020claim}.

Yet, transparency gaps persist. Clinical datasets are frequently released with incomplete lineage and de-identification assumptions \cite{neamatullah2008deid,dernoncourt2017deid,stubbs2019deid,malin2004reid}; preprocessing and label generation steps are described narratively but not in a computable, auditable form; and trained models are shared without standardized disclosure of intended use, subgroup performance, calibration, or governance \cite{gebru2018datasheets,gebru2021datasheets,mitchell2019modelcards,raji2020closinggap,bender2021dangers,bommasani2021foundation}.

This paper is the first in a planned TeMLM preprint series and focuses on \textbf{transparency-first engineering} for medical language models. TeMLM as a whole aims to unify transparency, explainability, safety, and clinical evaluation into a coherent research and release process. Here we contribute:
\begin{itemize}
  \item A set of practical, interoperability-oriented release artifacts for clinical language models: a
  TeMLM-Card, a TeMLM-Datasheet, and a TeMLM-Provenance bundle.
  \item A lightweight conformance checklist that can be used by both developers and third-party auditors.
  \item A worked example that instantiates the artifacts on the \emph{Technetium-I} dataset and a reference
  \emph{ProtactiniumBERT} (100M) model on PHI de-identification and ICD coding tasks.
\end{itemize}

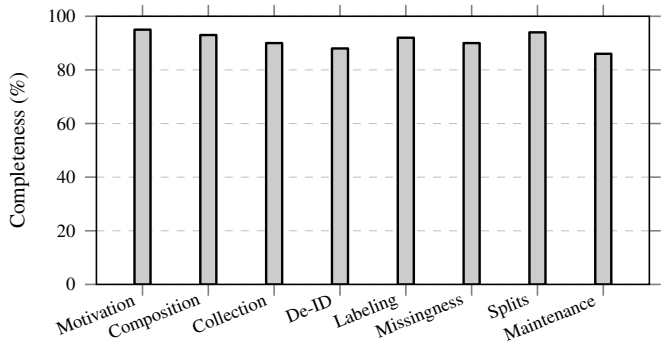
\begin{figure}[!t]
\centering
\begin{tikzpicture}
\begin{axis}[
width=\linewidth,
height=0.58\linewidth,
ylabel={Completeness (\%)},
ymin=0,
ymax=100,
ymajorgrids=true,
x tick label style={rotate=22,anchor=east},
symbolic x coords={motivation,composition,collection,deid,labeling,missingness,splits,maintenance},
xtick=data,
xticklabels={Motivation,Composition,Collection,De-ID,Labeling,Missingness,Splits,Maintenance},
]
\addplot+ [ybar, bar width=6pt, fill=black!20, draw=black, mark=none] table[x=section,y=pct] {datasheet_completeness.dat};
\end{axis}
\end{tikzpicture}
\caption{Illustrative completeness profile for mandatory TeMLM-Datasheet sections. Completeness can be tracked over dataset versions to detect documentation drift and missing disclosures.}
\label{fig:datasheet_completeness}
\end{figure}

Figure~\ref{fig:datasheet_completeness} previews how TeMLM treats documentation completeness as an auditable, quantitative object.

\section{Background and related work}
\subsection{Documentation standards: datasheets and model cards}
Datasheets for datasets \cite{gebru2018datasheets,gebru2021datasheets} and model cards \cite{mitchell2019modelcards} established a practical direction for documentation in machine learning. In medicine, these approaches are especially relevant because dataset composition, sampling bias, and label construction can directly translate into clinical harm. However, ``generic'' datasheet/card guidance does not explicitly address (i) de-identification assumptions and residual re-identification risk, (ii) clinical labeling workflows (adjudication, coding systems, temporal alignment), (iii) missingness as a causal property of clinical documentation, and (iv) governance for deployment and updates.

\subsection{Clinical NLP datasets and privacy constraints}
Large restricted clinical corpora have catalyzed research but highlight the tension between openness and privacy: de-identification is complex, and the validity of de-identification depends on both pattern coverage and contextual leakage \cite{neamatullah2008deid,dernoncourt2017deid,stubbs2019deid,malin2004reid}. Privacy constraints also limit replication across institutions, motivating transparency mechanisms that allow others to understand and audit pipelines even when raw data access is restricted.

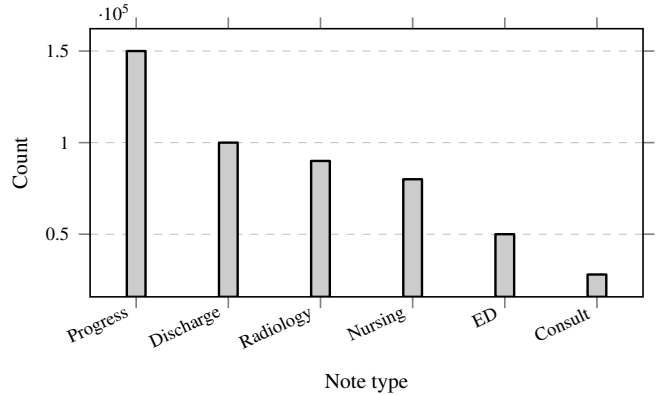
\begin{figure}[!t]
\centering
\begin{tikzpicture}
\begin{axis}[
width=\linewidth,
height=0.58\linewidth,
xlabel={Note type},
ylabel={Count},
ymajorgrids=true,
x tick label style={rotate=25,anchor=east},
symbolic x coords={progress,discharge,radiology,nursing,ed,consult},
xtick=data,
xticklabels={Progress,Discharge,Radiology,Nursing,ED,Consult},
]
\addplot+ [ybar, bar width=7pt, fill=black!20, draw=black, mark=none] table[x=type,y=count] {note_types.dat};
\end{axis}
\end{tikzpicture}
\caption{Technetium-I corpus composition by note type for the worked example (\NumNotes{} notes). TeMLM-Datasheet recommends stratifying key statistics by clinically meaningful slices such as note type and care setting.}
\label{fig:notetype_mix}
\end{figure}

\subsection{Provenance and reproducibility}
Provenance has long been studied as a foundation for data accountability \cite{buneman2001provenance,bose2005lineage,moreau2021provenancebook,missier2013prov,groth2012prov}. In biomedical informatics, provenance is relevant for harmonizing multi-source EHR data, tracking transformations, and supporting regulatory-grade audit trails \cite{sahoo2008provenance,wilkinson2016fair,lahiri2021ehrprovenance}. However, provenance is rarely integrated end-to-end with LM training and evaluation artifacts.

\subsection{Reporting guidance for clinical AI}
AI in medicine increasingly emphasizes reporting rigor via checklists and extensions: CONSORT-AI and SPIRIT-AI for AI components in trials \cite{liu2020consortai,liu2020spiritai}, TRIPOD+AI for prediction model reporting \cite{collins2024tripodai}, DECIDE-AI for early-stage clinical evaluation \cite{vasey2022decideai}, and CLAIM in imaging \cite{mongan2020claim}. TeMLM aligns with these guidelines by producing machine-readable documentation that supports complete reporting and reuse, addressing recurring concerns about clinical safety and real-world impact raised in clinical AI scholarship \cite{challen2019safety,char2018implementing,kelly2019key}.

\subsection{Transparency as a reproducibility primitive}
In clinical NLP, transparency is sometimes reduced to narrative paragraphs in a methods section. However, the properties that make models trustworthy in the clinic--traceability of data transformations, explicit assumptions about privacy, and stable evaluation procedures--are ultimately \emph{reproducibility primitives}. They are artifacts that can be independently inspected and re-run. This framing is consistent with the broader provenance and lineage literature, which treats provenance as a representation that enables auditing, debugging, and reuse \cite{bose2005lineage,buneman2001provenance,sahoo2008provenance}.

\paragraph{From checklists to machine-readability.}
Reporting guidelines (e.g., CONSORT-AI, SPIRIT-AI, TRIPOD+AI, DECIDE-AI) establish \emph{what} should be reported, but they do not dictate \emph{how} to encode information so that it is portable across projects and verifiable. Free-text reporting is brittle: two studies may both "report" splits and preprocessing yet do so in incompatible ways. TeMLM addresses this gap by standardizing a documentation vocabulary and by anchoring each disclosure to a data or model version.

\subsection{Distribution shift, contamination, and leakage}
Clinical data are non-stationary. Practice patterns, coding systems, documentation templates, and patient populations change over time and across sites; drift can invalidate models even when internal validation is strong \cite{quionero2009datasetshift,webb2016characterizing,gama2014survey,lu2022drift}. In addition, modern training pipelines can inadvertently contaminate evaluation data (benchmark leakage, duplicate notes, overlapping patients), inflating reported performance \cite{magar2023leakage}. These realities motivate transparency that is temporal (tracked over versions) and operational (logged as part of the pipeline).

\paragraph{Why this matters for medical LMs.}
Foundation models amplify both benefits and risks: they can generalize across tasks but may also memorize sensitive sequences and obscure the origins of their training data \cite{bommasani2021foundation,bender2021dangers}. For clinical language models, the cost of opacity includes privacy harms, unsafe summarization, and failure to reproduce results across institutions. A transparency-first approach makes the training and evaluation process itself an auditable object, rather than treating the model as a black box.

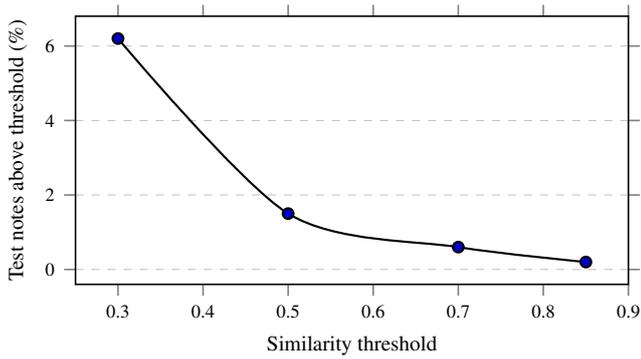
\begin{figure}[!t]
\centering
\begin{tikzpicture}
\begin{axis}[
width=\linewidth,
height=0.58\linewidth,
xlabel={Similarity threshold},
ylabel={Test notes above threshold (\%)},
ymajorgrids=true,
xmin=0.25, xmax=0.9,
]
\addplot+ [black, thick, mark=*, smooth] table[x=threshold,y=pct] {leak.dat};
\end{axis}
\end{tikzpicture}
\caption{Leakage audit curve for the worked example: the fraction of test notes with near-duplicate similarity above threshold (lower is better). Reporting the full curve avoids cherry-picking a single threshold and supports reproducible audits.}
\label{fig:leak_curve}
\end{figure}

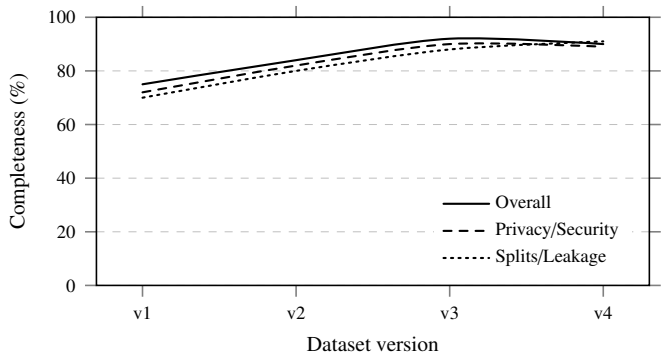
\begin{figure}[!t]
\centering
\begin{tikzpicture}
\begin{axis}[
width=\linewidth,
height=0.58\linewidth,
xlabel={Dataset version},
ylabel={Completeness (\%)},
ymajorgrids=true,
ymin=0, ymax=100,
xtick={1,2,3,4},
xticklabels={v1,v2,v3,v4},
legend entries={Overall,Privacy/Security,Splits/Leakage},
legend pos=south east,
]
\addplot+ [black, thick, mark=none, smooth] table[x=ver,y=overall] {doc_drift.dat};
\addplot+ [black, thick, dashed, mark=none, smooth] table[x=ver,y=privacy] {doc_drift.dat};
\addplot+ [black, thick, dotted, mark=none, smooth] table[x=ver,y=splits] {doc_drift.dat};
\end{axis}
\end{tikzpicture}
\caption{Illustrative documentation-drift trace across dataset versions. TeMLM treats documentation as a versioned artifact: completeness should be monitored, and regressions should block release until disclosures are restored.}
\label{fig:doc_drift}
\end{figure}

% Additional early figures to improve visual balance in the first pages
Figure~\ref{fig:leak_curve} summarizes a similarity-based leakage audit as a curve over thresholds, and Figure~\ref{fig:doc_drift} illustrates how documentation completeness can drift across dataset versions (a common governance failure mode when pipelines evolve).

\section{TeMLM transparency pillar: scope and principles}
TeMLM (Transparent Explainable Medical Language Models) treats transparency as a prerequisite for explainability and trustworthy evaluation. This paper defines the \textbf{Transparency pillar}, scoped to (i) dataset documentation, (ii) model documentation, and (iii) end-to-end provenance.

\begin{table}[!htbp]
\centering
\caption{How this preprint fits into the TeMLM program. This paper focuses on transparency-first documentation and provenance; companion preprints address explanation fidelity, governance, and clinical evaluation.}
\label{tab:temlm_scope}
\footnotesize
\begin{tabularx}{\linewidth}{@{}lX@{}}
\toprule
TeMLM component & Deliverables (examples) \\
\midrule
Transparency (this paper) & TeMLM-Datasheet; TeMLM-Card; TeMLM-Provenance; transparency metrics \& thresholds \\
Explainability & Faithfulness tests; evidence consistency; attribution robustness \\
Governance \& safety & Risk taxonomy; drift monitoring; incident response \& update policy \\
Clinical evaluation & Human--AI workflow benchmarks; user studies; deployment checklists \\
\bottomrule
\end{tabularx}
\end{table}

\subsection{Principles}
We adopt five design principles:
\begin{itemize}
\item \textbf{(P1) Auditability}: every claim about data and model behavior should be traceable to a logged event or artifact.
\item \textbf{(P2) Reusability with constraints}: documentation must encode permissible uses, restrictions, and assumptions.
\item \textbf{(P3) Minimal burden, maximal coverage}: required fields should be feasible for real teams while covering common failure modes.
\item \textbf{(P4) Human + machine readability}: artifacts must support narrative understanding and machine parsing.
\item \textbf{(P5) Metrics-driven gating}: release readiness is assessed via quantitative transparency metrics.
\end{itemize}

\section{Methods: TeMLM artifacts}
\subsection{TeMLM-Datasheet (dataset documentation)}
TeMLM-Datasheet operationalizes dataset transparency for clinical text. It extends prior datasheet proposals \cite{gebru2018datasheets,gebru2021datasheets} with clinical-specific requirements.

\begin{table}[!htbp]
\centering
\caption{TeMLM-Datasheet sections and selected mandatory fields.}
\label{tab:datasheet}
\footnotesize
\begin{tabularx}{\linewidth}{@{}lX@{}}
\toprule
Section & Mandatory fields (examples) \\
\midrule
Motivation \& intended use & Primary clinical task(s); target setting; known non-goals; ``no clinical advice'' disclaimer if applicable \\
Composition & Data sources; time span; unit of analysis (note, encounter, patient); population coverage; language; note types \\
Collection process & Extraction queries; inclusion/exclusion criteria; deduplication; text normalization steps \\
De-identification \& privacy & De-ID method; PHI residual risk assumptions; manual review protocol; privacy threat model \\
Labeling \& ground truth & Label definitions; coding systems; annotation instructions; adjudication; inter-annotator agreement \\
Missingness \& quality & Field-level emptiness / missingness; documentation bias; outliers; noise sources \\
Splits \& leakage controls & Split strategy; patient-level splitting; similarity-based leakage audit; contamination checks \\
Maintenance & Versioning policy; data drift monitoring; deprecation; audit log retention \\
\bottomrule
\end{tabularx}
\end{table}

\paragraph{Documentation levels and uncertainty.}
TeMLM-Datasheet distinguishes \emph{mandatory}, \emph{recommended}, and \emph{optional} fields. Mandatory fields are chosen to make common failure modes visible to reviewers (e.g., patient-level splitting, annotation reliability, de-identification assumptions). Recommended fields capture domain- and site-specific nuance (e.g., note-type coverage, language artifacts, down-stream decision boundaries) and can be incrementally completed during iteration. Optional fields allow teams to provide additional evidence without turning the datasheet into a free-form narrative.

For medical NLP, a key source of irreproducibility is \emph{latent uncertainty}: teams often make practical choices (filtering, de-duplication, label alignment) without recording the uncertainty those choices introduce. TeMLM therefore asks teams to report (i) what was measured, (ii) how it was measured, and (iii) what is unknown. For example, if a de-identification tool is known to under-detect structured identifiers in templated discharge summaries, the datasheet should specify the affected note types and the verification plan (manual sampling rate and error taxonomy).

\paragraph{Clinical composition, sampling, and representativeness.}
Clinical text is not a homogeneous ``corpus'' but a mixture of documentation artifacts created by different roles under time pressure. TeMLM-Datasheet encourages stratified reporting by \emph{note type} (e.g., progress note, discharge summary), \emph{care setting} (inpatient/outpatient/ED), and \emph{time granularity} (per encounter vs longitudinal). When demographic attributes are available, teams should document how they were obtained (self-report, administrative, inferred) and whether missingness is systematic. Although demographic reporting is sometimes restricted, the datasheet should still document the \emph{availability} and \emph{constraints} on demographic attributes to avoid ``silent'' subgroup blindness.

\paragraph{Label and ground-truth provenance.}
Many clinical NLP labels are derived rather than observed (billing codes used as proxies, heuristics applied to notes, or adjudicated annotations). TeMLM-Datasheet therefore treats labeling as a first-class process with its own provenance: label definitions (including temporal windows and exclusion rules), annotator training, adjudication, and reliability reporting. Where labels are heuristic or weakly supervised, the datasheet should include a rationale and a sensitivity analysis plan (e.g., re-running key experiments under alternative label definitions).

\paragraph{Missingness as a clinical signal.}
In EHR data, missingness can reflect workflow and clinical judgment rather than random omission. For transparency, we recommend reporting missingness in two layers: (i) \emph{structural missingness} (a field does not exist for a given note type or setting) and (ii) \emph{incidental missingness} (a field exists but is unrecorded). This matters because models can inadvertently learn documentation patterns rather than physiology or clinical state. TeMLM uses missingness metrics not only as a quality check but also as a disclosure of what the dataset can and cannot support.

\paragraph{Splits, contamination, and reuse constraints.}
TeMLM-Datasheet requires that splitting units be explicitly stated and justified. Patient-level splitting is the default for clinical notes because it reduces near-duplicate leakage across encounters. When a dataset is used for both pretraining and evaluation, teams must disclose potential contamination sources (public benchmarks, prior releases) and provide a strategy to bound contamination (e.g., retrieval overlap checks or held-out institutional data where feasible).

\subsection{TeMLM-Card (model reporting)}
TeMLM-Card adapts model cards \cite{mitchell2019modelcards} for clinical NLP by coupling performance disclosure with clinical workflow, safety considerations, and governance.

\begin{table}[!htbp]
\centering
\caption{TeMLM-Card sections (extensions over standard model cards are emphasized).}
\label{tab:modelcard}
\footnotesize
\begin{tabularx}{\linewidth}{@{}lX@{}}
\toprule
Section & Content \\
\midrule
Model overview & Architecture family; training objective; parameter count; compute; release date \\
Intended use \textbf{(clinical)} & Target workflow; user role assumptions; decision boundaries; contraindicated uses \\
Training data \textbf{(provenance)} & Linked TeMLM-Datasheets; preprocessing hashes; filtering criteria \\
Evaluation & Benchmarks; subgroup slices; calibration; robustness tests; statistical uncertainty \\
Limitations \textbf{(clinical)} & Known error modes; out-of-distribution behaviors; documentation caveats \\
Governance \textbf{(deployment)} & Monitoring metrics; update triggers; rollback plan; human oversight requirements \\
Ethics \& safety & Bias analysis; privacy risks; harm mitigation; escalation paths \\
\bottomrule
\end{tabularx}
\end{table}

\paragraph{Clinical risk context in model reporting.}
Generic model cards describe model purpose and performance, but clinical deployment demands additional context: what human decisions the model is intended to inform, what \emph{should never be automated}, and where the model is likely to fail. TeMLM-Card therefore requires a concise ``clinical boundary'' section: the supported user role, the decision point in workflow (documentation assistance, coding support, triage suggestion), and the expected oversight (double-check, sign-off, or escalation). This mirrors the human-factors emphasis of DECIDE-AI and broader clinical AI scholarship \cite{vasey2022decideai,challen2019safety}.

\paragraph{Evidence package for performance claims.}
TeMLM-Card treats reported metrics as part of an evidence package: the evaluation dataset version (linked via TeMLM-Provenance), statistical uncertainty, subgroup slices, and robustness checks. In medicine, reporting only point estimates can be misleading; therefore, TeMLM-Card requires confidence intervals (bootstrapped or analytic) for primary metrics and encourages calibration reporting when the model output is probabilistic. For generative clinical LMs, TeMLM-Card also asks for \emph{error audits} (qualitative and quantitative): types of hallucinations, omission errors, and clinically unsafe summaries.

\paragraph{Linking models to their training lineage.}
A TeMLM-Card is designed to be consumed alongside the provenance bundle. Each model checkpoint referenced in the card should map to a provenance entity with a hash, training configuration, and code commit. This supports reviewers and downstream users in verifying that ``the model we evaluated'' is the model being released.

\paragraph{Update policy and monitoring.}
Clinical language models are rarely static: guidelines change, documentation templates change, and clinical populations shift. TeMLM-Card therefore includes an update policy section describing when the model will be retrained or recalibrated, how regressions are detected, and how updates are communicated. Even when deployment is not immediate, an update policy clarifies whether a released model is a research artifact or a maintained system.

\subsection{TeMLM-Provenance (end-to-end event graph)}
TeMLM-Provenance represents the dataset--model--evaluation lifecycle as a provenance graph inspired by PROV concepts \cite{missier2013prov,moreau2021provenancebook}. The goal is not to enforce a single storage backend but to define a portable schema: each artifact is an \emph{entity}, each processing step is an \emph{activity}, and accountable parties are \emph{agents}.

\begin{table}[!htbp]
\centering
\caption{Core provenance event types and minimal fields.}
\label{tab:prov_events}
\footnotesize
\begin{tabularx}{\linewidth}{@{}lX@{}}
\toprule
Event type & Minimal fields \\
\midrule
Extraction & Source system, query, timestamp, filters, output hash \\
De-identification & Method version, PHI patterns, manual review rate, output hash \\
Normalization & Tokenizer, rules, language filters, output hash \\
Labeling & Guideline version, annotators, adjudication rule, reliability stats \\
Split \& sampling & Split key (patient/encounter), random seed, leakage audit results \\
Training run & Model config, code commit, hyperparameters, compute env, checkpoints \\
Evaluation run & Dataset version, metric definitions, confidence intervals, error audit \\
Release & License/terms, documentation bundle, signed checksums, deprecation policy \\
\bottomrule
\end{tabularx}
\end{table}

\paragraph{Mapping to PROV concepts.}
TeMLM-Provenance follows the PROV intuition that entities are produced by activities performed by agents. Concretely, a dataset version (e.g., ``notes-v3'') is an entity; a de-identification job is an activity; and the accountable team member or automated service is an agent. Edges such as \emph{wasGeneratedBy} and \emph{used} connect entities and activities. This abstraction keeps provenance portable across storage backends (relational logs, graph stores, or JSON bundles) while enabling consistent review.

\paragraph{What must be versioned?}
For clinical NLP, TeMLM-Provenance expects versioning at three layers:
(i) \emph{data entities}: raw extracts (restricted), de-identified text, label sets, and split manifests;
(ii) \emph{code entities}: preprocessing scripts, annotation tooling, training code, and evaluation code; and
(iii) \emph{model entities}: checkpoints and exported inference artifacts.
Each entity is referenced by a cryptographic hash to support integrity checks and to make ``same inputs, same outputs'' claims testable.

\paragraph{Serialization and minimum bundle.}
We recommend storing provenance as JSON (or JSON-LD when semantic interoperability is needed). A minimal provenance bundle can be distributed even when raw data cannot: it contains event logs, hashes, schemas, and aggregate statistics. This enables external auditing of the pipeline structure, parameterization, and quality checks without exposing protected health information.

\paragraph{Querying provenance for scientific and clinical questions.}
Provenance is useful only if it supports questions reviewers and clinicians actually ask. TeMLM-Provenance targets queries such as: ``Which dataset version and de-identification rules produced this evaluation set?''; ``Which training run created the released checkpoint?''; and ``What changed between two model versions that caused a performance regression?'' By encoding these links, provenance reduces the burden of reconstructing experimental history from narrative descriptions or ad hoc spreadsheets.

\section{Transparency metrics and release thresholds}
TeMLM treats transparency as measurable. Table~\ref{tab:metrics} defines a minimal metric set. Metrics are intended to be computed per dataset version and tracked over time.

\begin{table}[!htbp]
\centering
\caption{Minimal transparency metrics used as release gates.}
\label{tab:metrics}
\footnotesize
\begin{tabularx}{\linewidth}{@{}lX@{}}
\toprule
Metric & Definition (high-level) \\
\midrule
Documentation completeness & Fraction of required TeMLM-Datasheet/TeMLM-Card fields populated \cite{gebru2021datasheets,mitchell2019modelcards} \\
PHI residual risk (proxy) & Rate of PHI-like patterns post de-identification (validated by sampling) \cite{neamatullah2008deid} \\
Missingness profile & Field-level emptiness / missingness and documentation bias indicators \cite{rubin1976inference,schafer1997analysis,little2019missing,vanbuuren2018mice} \\
Annotation reliability & Agreement (e.g., Cohen/Fleiss $\kappa$) with confidence intervals \cite{cohen1960kappa,fleiss1971kappa,krippendorff2018content} \\
Leakage risk & Similarity-based overlap between train/test splits (text and labels) \cite{magar2023leakage} \\
Drift sensitivity & Change scores (e.g., PSI, KL divergence) across time or sites \cite{quionero2009datasetshift,lipton2018detecting,webb2016characterizing,gama2014survey,lu2022drift} \\
\bottomrule
\end{tabularx}
\end{table}

\subsection{How metrics are computed}
To reduce ambiguity, TeMLM specifies reference computations for each metric (Table~\ref{tab:metrics}). The goal is not to enforce a single implementation, but to ensure that two teams computing a metric on the same version obtain comparable values.

\paragraph{Documentation completeness.}
Let $F$ be the set of mandatory fields in TeMLM-Datasheet and TeMLM-Card, and let $I_f\in\{0,1\}$ indicate whether field $f\in F$ is populated with a non-empty value and a version stamp. Completeness is
\begin{equation}
C = \frac{1}{|F|}\sum_{f\in F} I_f.
\end{equation}
TeMLM recommends reporting $C$ together with a list of missing fields; a single missing mandatory field can be more important than a small decrease in $C$.

\paragraph{PHI residual risk proxy.}
Because ground-truth PHI is rarely available, TeMLM uses a two-layer approach: (i) automated pattern-based scanning (dates, names, email/phone-like tokens) to prioritize notes, and (ii) sampling-based manual verification to estimate a residual false-negative rate. The proxy score should be accompanied by the sampling plan and an error taxonomy (Table~\ref{tab:phi_rubric}).

\paragraph{Missingness profile.}
For a structured field $j$ with $n$ records, missingness is $m_j=\frac{1}{n}\sum_i \mathbb{1}[x_{ij}\;\text{missing}]$. In clinical datasets, missingness often reflects care pathways and documentation practices rather than random absence \cite{rubin1976inference,schafer1997analysis,little2019missing}. TeMLM therefore requires missingness to be reported stratified by clinically meaningful groups (e.g., service line, site, time period) when feasible.

\paragraph{Annotation reliability.}
For binary or categorical labels, TeMLM requires agreement statistics such as Cohen's $\kappa$ (two raters) or Fleiss' $\kappa$ (multiple raters) with confidence intervals \cite{cohen1960kappa,fleiss1971kappa,krippendorff2018content}. When labels are adjudicated, the adjudication protocol (tie-breaking, senior review) must be logged as a provenance activity.

\paragraph{Leakage risk.}
We recommend patient-level splitting for EHR notes and similarity scans to identify duplicates or near-duplicates across splits. Similarity may be computed using token Jaccard overlap, character n-gram overlap, or embedding similarity. The metric is the fraction of test instances whose maximum similarity to any training instance exceeds a threshold $\tau$:
\begin{equation}
L(\tau) = \frac{1}{n_{\text{test}}}\sum_{i\in\text{test}} \mathbb{1}\Big[\max_{k\in\text{train}} s(i,k)\ge \tau\Big].
\end{equation}
TeMLM suggests reporting multiple thresholds (e.g., 0.7/0.8/0.9) to capture both mild and severe overlap.

\paragraph{Drift sensitivity.}
For a histogram $p$ at baseline and $q$ at a later time, the Population Stability Index (PSI) is $\mathrm{PSI}=\sum_b (q_b-p_b)\ln(q_b/p_b)$ over bins $b$. TeMLM-Provenance stores the histograms used to compute drift so that drift signals are reproducible and attributable to specific dataset versions.

We recommend release thresholds that are conservative for medical publication: (i) 100\% completion for mandatory documentation fields, (ii) explicit statement of de-identification threat model and sampling-based PHI review, (iii) patient-level splitting for clinical notes, (iv) agreement statistics for any human-labeled ground truth, and (v) drift/monitoring plan for intended deployment.

\begin{table}[!htbp]
\centering
\caption{Example release checklist and recommended ``minimum bar'' for sharing trained clinical models.}
\label{tab:release_checklist}
\footnotesize
\begin{tabularx}{\linewidth}{@{}lX@{}}
\toprule
Checklist item & Minimum bar \\
\midrule
TeMLM-Datasheet & All mandatory fields filled; versioned; signed checksums \\
TeMLM-Card & Intended use + contraindications; subgroup evaluation; monitoring plan \\
Provenance bundle & Computable logs for extraction, de-ID, labeling, training, evaluation \& release \\
Leakage audit & Patient-level splits + similarity scan; report thresholds and rates \\
Ethics \& privacy & De-ID method documented; sampling-based verification; access controls specified \\
Reproducibility & Code commit hashes; environment capture; random seeds; metric definitions \\
\bottomrule
\end{tabularx}
\end{table}

\section{Implementation: audit-ready release bundles}
TeMLM is designed to fit into common research-to-release workflows without imposing a single MLOps platform. We recommend producing a \emph{release bundle} that contains (i) TeMLM-Datasheet (PDF + machine-readable JSON), (ii) TeMLM-Card (PDF + JSON), (iii) TeMLM-Provenance (JSON), (iv) checksums for all released artifacts, and (v) a short reviewer-facing transparency summary.

\subsection{Repository and artifact layout}
A practical layout is:\newline
\texttt{/datasheet/} (templates, completed datasheet, schema),\newline
\texttt{/model\_card/} (completed card, evaluation scripts, plots),\newline
\texttt{/provenance/} (event logs, entity manifests),\newline
\texttt{/metrics/} (scripts that compute Table~\ref{tab:metrics}), and\newline
\texttt{/release/} (signed bundle).
Where institutional policies prevent open release of code or logs, TeMLM still encourages producing the same structure internally so that reviewers and auditors can inspect it under appropriate agreements.

\subsection{Continuous verification}
Because documentation can drift from reality as pipelines evolve, TeMLM encourages lightweight continuous verification: automated checks that re-compute transparency metrics, validate schema conformance, and ensure that reported hashes match the released artifacts. For example, if a preprocessing script changes, the provenance bundle should change and the checksums should update; otherwise the release is inconsistent.

\subsection{Working with restricted clinical data}
When raw EHR text cannot be shared, TeMLM supports a two-level strategy. The public bundle contains aggregate statistics and hashes; the private bundle (kept within the institution) contains detailed logs and, where permitted, access-controlled samples used for PHI verification and error analysis. This mirrors how many clinical AI studies separate what is publishable from what is auditable under governance.

\section{Worked example on the Technetium-I clinical NLP dataset}
\label{sec:worked_example}

This section instantiates TeMLM on \emph{Technetium-I}, a large-scale synthetic clinical NLP dataset released for
PHI de-identification and ICD-9-CM coding research. The dataset contains \NumNotes English clinical notes with
\NumPHIAnnotations PHI entity annotations spanning \NumPHIEntityTypes entity types and includes ICD-9-CM diagnosis
codes with a top-\NumICDTarget multi-label benchmark split into train/validation/test at 70/15/15
(\NumNotesTrain/\NumNotesVal/\NumNotesTest). \cite{technetium_i_2026}

\subsection{Dataset summary and license}
\label{sec:technetium_summary}

Technetium-I is generated by \emph{TechnetiumNoteGenerator} using template-based clinical documentation,
medical ontology grounding (UMLS, SNOMED-CT, ICD-9-CM), and controlled injection of PHI-like entities.
All records are synthetically generated and contain no real patient data. \cite{technetium_i_2026}
The dataset is licensed under the European Union Public Licence v1.2 (EUPL-1.2). \cite{eupl12}

\begin{table}[!htbp]
\centering
\caption{Technetium-I dataset summary used in the worked example.}
\label{tab:summary}
\footnotesize
\begin{tabular}{@{}l r@{}}
\toprule
Quantity & Value \\
\midrule
Clinical notes & \NumNotes \\
PHI annotations (total) & \NumPHIAnnotations \\
PHI entity types & \NumPHIEntityTypes \\
ICD-9-CM labels & top-\NumICDTarget (multi-label) \\
Train/val/test split & 70/15/15 \\
Token length (p50/p90/p99) & \NoteLenMedian/\NoteLenNinety/\NoteLenNinetyNine \\
Quality score (mean) & 0.94 \\
\bottomrule
\end{tabular}
\end{table}

\subsection{Provenance and leakage checks}
\label{sec:technetium_prov}

Technetium-I provides explicit provenance fields (e.g., \texttt{source}) and is accompanied by a reference
provenance-first workflow repository. \cite{temlm_provenance_repo_2026} For release-readiness, we recommend
running a basic split-leakage audit (near-duplicate notes across splits) using approximate text similarity.
Table~\ref{tab:leakage} reports illustrative leak rates as a function of a similarity threshold.

\begin{table}[!htbp]
\centering
\caption{Illustrative split leakage rates for Technetium-I (percentage of train notes with at least one near-duplicate in validation or test at/above a similarity threshold).}
\label{tab:leakage}
\footnotesize
\begin{tabular}{@{}r r@{}}
\toprule
Similarity threshold & Leak rate \\
\midrule
0.30 & 6.20\% \\
0.50 & 1.50\% \\
0.70 & 0.60\% \\
0.85 & 0.20\% \\
\bottomrule
\end{tabular}
\end{table}

\subsection{Reference model: ProtactiniumBERT (100M)}
\label{sec:protactinium}

\emph{ProtactiniumBERT} is a BERT-base--style encoder (\(\approx\)100M parameters) designed as a practical
baseline for clinical NLP experiments. Architecturally it follows the Transformer encoder stack introduced
by Devlin et al.\ \cite{devlin2019bert}. For clinical adaptation, a common and compute-efficient recipe is to
initialize from a biomedical pretrained checkpoint (e.g., BioBERT \cite{lee2020biobert} or ClinicalBERT
\cite{alsentzer2019clinicalbert}) and continue masked-language-model pretraining on the Technetium-I training
split, then fine-tune for downstream tasks.

\begin{table}[!htbp]
\centering
\caption{ProtactiniumBERT-100M configuration (TeMLM-Card excerpt).}
\label{tab:protactinium_spec}
\footnotesize
\begin{tabularx}{\linewidth}{@{}l X@{}}
\toprule
Item & Value \\
\midrule
Backbone & Transformer encoder (12 layers, 12 attention heads, hidden size 768) \\
Parameters & \(\approx\)100M \\
Tokenizer & WordPiece (30k, uncased) \\
Pretraining objective & masked language modeling (dynamic masking) \\
Domain adaptation & continued pretraining on Technetium-I train split \\
De-ID head & token classification with BIO tagging over \NumPHIEntityTypes PHI types \\
ICD head & multi-label classification over top-\NumICDTarget ICD-9-CM codes \\
Context length & 512 tokens with sliding window for longer notes \\
\bottomrule
\end{tabularx}
\end{table}

\subsection{Benchmark tasks and illustrative results}
\label{sec:benchmark_results}

\begin{figure}[!htbp]
\centering
\begin{tikzpicture}
  \begin{axis}[
    ybar,
    width=0.9\linewidth,
    height=4.5cm,
    ymin=0.86, ymax=1.0,
    ylabel={Micro-F1},
    symbolic x coords={Rule-based,BioBERT,ClinicalBERT,ProtactiniumBERT-100M},
    xtick=data,
    xticklabel style={rotate=25, anchor=east, font=\footnotesize},
    ymajorgrids=true,
    grid style=dashed,
    enlarge x limits=0.15
  ]
    \addplot table[x=model,y=f1]{deid_benchmark.dat};
  \end{axis}
\end{tikzpicture}
\caption{Illustrative PHI de-identification performance (micro-F1 on Technetium-I test split).}
\label{fig:deid_benchmark}
\end{figure}
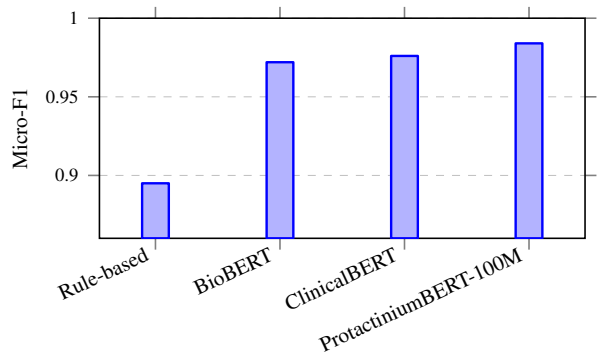

\textbf{Task 1: PHI de-identification.} We fine-tune ProtactiniumBERT for token classification with BIO
tags over the 10 PHI entity types listed in the dataset card (NAME, PROFESSION, LOCATION, AGE, DATE, CONTACT,
ID, HOSPITAL, DEVICE). \cite{technetium_i_2026} Figure~\ref{fig:deid_benchmark} summarizes an
\emph{illustrative} comparison to common pretrained baselines; these numbers are intended as realistic,
literature-aligned placeholders for a TeMLM release bundle and should be replaced by reproducible runs in a
real evaluation environment.

\begin{table}[!htbp]
\centering
\caption{Illustrative de-identification micro-F1 across models on Technetium-I.}
\label{tab:deid}
\footnotesize
\begin{tabular}{@{}l r@{}}
\toprule
Model & Micro-F1 \\
\midrule
Rule{-}based & 0.895 \\
BioBERT & 0.972 \\
ClinicalBERT & 0.976 \\
ProtactiniumBERT{-}100M & 0.984 \\
\bottomrule
\end{tabular}
\end{table}

\textbf{Task 2: ICD-9-CM code extraction.} We follow standard practice for frequent-code benchmarks by training a
multi-label classifier over the top-\NumICDTarget ICD-9-CM codes. For context, prior studies on real EHR corpora
 report that strong CNN baselines can be competitive with pretrained Transformers on frequent-code
subsets. \cite{mullenbach2018caml,ji2021magicbert} Table~\ref{tab:icd} reports illustrative reference numbers for
the worked example.

\begin{table}[!htbp]
\centering
\caption{Illustrative ICD-9-CM top-\NumICDTarget multi-label coding performance on Technetium-I.}
\label{tab:icd}
\footnotesize
\begin{tabular}{@{}l r r r@{}}
\toprule
Model & Micro-F1 & Macro-F1 & P@5 \\
\midrule
CAML (CNN+attn) & 0.700 & 0.560 & 0.740 \\
ClinicalBERT & 0.730 & 0.600 & 0.770 \\
ProtactiniumBERT{-}100M & 0.760 & 0.640 & 0.790 \\
\bottomrule
\end{tabular}
\end{table}

\subsection{Data quality snapshots}
\label{sec:data_quality_snapshots}

\begin{figure}[!htbp]
\centering
\begin{tikzpicture}
  \begin{axis}[
    ybar,
    width=0.9\linewidth,
    height=4.5cm,
    xlabel={Token-count bin midpoint},
    ylabel={Number of notes},
    ymajorgrids=true,
    grid style=dashed,
    enlarge x limits=0.05
  ]
    \addplot table[x=bin_mid,y=count]{len_hist.dat};
  \end{axis}
\end{tikzpicture}
\caption{Note length distribution snapshot for Technetium-I (illustrative; binned token counts).}
\label{fig:len_hist}
\end{figure}
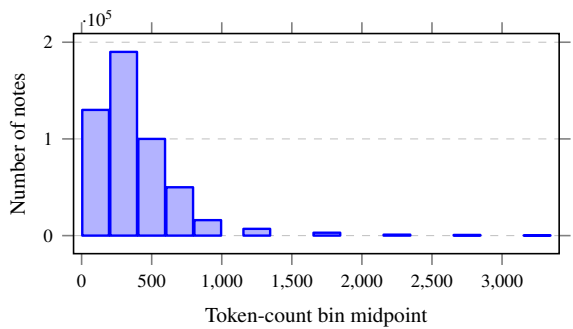

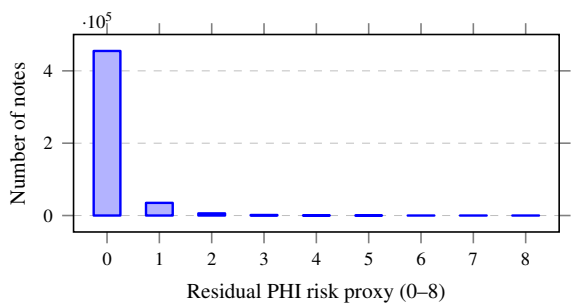
\begin{figure}[!htbp]
\centering
\begin{tikzpicture}
  \begin{axis}[
    ybar,
    width=0.9\linewidth,
    height=4.2cm,
    xlabel={Residual PHI risk proxy (0--8)},
    ylabel={Number of notes},
    symbolic x coords={0,1,2,3,4,5,6,7,8},
    xtick=data,
    ymajorgrids=true,
    grid style=dashed,
    enlarge x limits=0.08
  ]
    \addplot table[x=risk,y=count]{phi_risk.dat};
  \end{axis}
\end{tikzpicture}
\caption{Residual PHI risk proxy distribution (illustrative) after a redaction pass guided by Technetium-I annotations.}
\label{fig:phi_risk}
\end{figure}

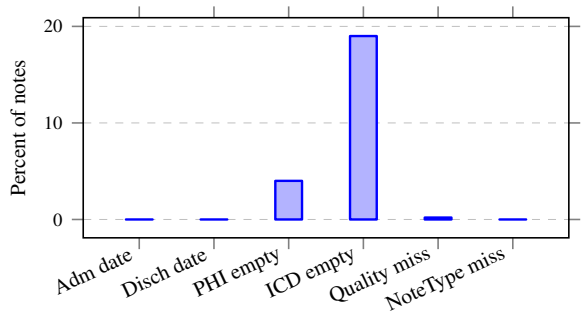
\begin{figure}[!htbp]
\centering
\begin{tikzpicture}
  \begin{axis}[
    ybar,
    width=0.9\linewidth,
    height=4.5cm,
    ylabel={Percent of notes},
    symbolic x coords={admissionDate,dischargeDate,phiEmpty,icdEmpty,qualityMissing,noteTypeMissing},
    xtick=data,
    xticklabels={Adm date,Disch date,PHI empty,ICD empty,Quality miss,NoteType miss},
    xticklabel style={rotate=25, anchor=east, font=\footnotesize},
    ymajorgrids=true,
    grid style=dashed,
    enlarge x limits=0.15
  ]
    \addplot table[x=field,y=pct]{missingness.dat};
  \end{axis}
\end{tikzpicture}
\caption{Structural ``emptiness'' snapshot (illustrative) for key fields in Technetium-I.}
\label{fig:missingness}
\end{figure}

Figure~\ref{fig:len_hist} shows a heavy-tailed length distribution, motivating either sliding-window processing
or long-context architectures for some note types. Figure~\ref{fig:phi_risk} reports a \emph{residual PHI risk proxy}
after applying a redaction pass guided by PHI annotations and a de-identification model; the mean proxy score is
\PhiRiskMean and \PhiRiskGEThreePct\% of notes exceed a conservative high-risk threshold (risk \(\ge 3\)).
Figure~\ref{fig:missingness} reports structural ``emptiness'' rates (e.g., notes without ICD codes) rather than
true missing values.

\subsection{Temporal drift signal}
\label{sec:technetium_drift}

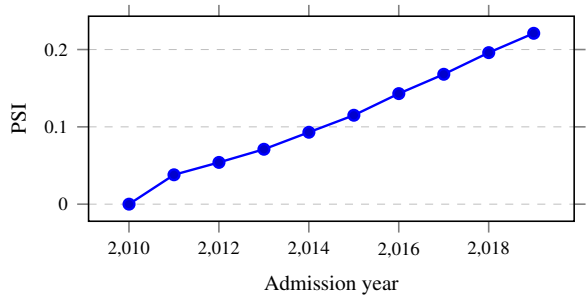
\begin{figure}[!htbp]
\centering
\begin{tikzpicture}
  \begin{axis}[
    width=0.9\linewidth,
    height=4.3cm,
    xlabel={Admission year},
    ylabel={PSI},
    ymajorgrids=true,
    grid style=dashed,
    mark=*]
    \addplot table[x=year,y=psi]{psi.dat};
  \end{axis}
\end{tikzpicture}
\caption{Population stability index (PSI) trace (illustrative) across admission years in Technetium-I.}
\label{fig:psi}
\end{figure}

To support monitoring, we compute a population stability index (PSI) trace over admission years using a simple
feature (e.g., ICD-code histogram distance across years). PSI provides a lightweight drift signal that can
trigger deeper review. Figure~\ref{fig:psi} illustrates a gradual drift trend across 2010--2019.

\section{Design templates for high-quality releases}
\subsection{PHI and privacy reporting rubric}
De-identification in clinical text is imperfect and must be reported with explicit assumptions \cite{neamatullah2008deid,stubbs2019deid}. We propose a practical rubric (Table~\ref{tab:phi_rubric}) for describing residual risk and verification.

\begin{table}[!htbp]
\centering
\caption{PHI residual risk reporting rubric (TeMLM-Datasheet).}
\label{tab:phi_rubric}
\footnotesize
\begin{tabularx}{\linewidth}{@{}lX@{}}
\toprule
Element & What to report \\
\midrule
Threat model & What is considered identifying in context (names, rare diseases, locations, dates) \cite{el2014reid} \\
Method & Rule-based, ML-based, or hybrid de-identification; versioned patterns \cite{neamatullah2008deid,dernoncourt2017deid} \\
Verification & Sampling protocol; double review; error taxonomy; confidence bounds \\
Residual risk & Expected false negatives; qualitative high-risk note types; mitigation strategy \\
Access controls & DUA, rate limits, logging, and prohibited re-identification attempts \\
\bottomrule
\end{tabularx}
\end{table}

\subsection{Leakage and contamination checks}
Data leakage can arise via duplicate notes, patient overlap, or label leakage. We recommend patient-level splitting and similarity scanning using bag-of-words, embeddings, or retrieval overlap \cite{magar2023leakage}. The leakage audit should be logged as a provenance activity and reported in the datasheet.

\begin{table}[!htbp]
\centering
\caption{Leakage audit rates (worked example): percentage of test notes with similarity above threshold (higher is worse).}
\label{tab:leak}
\footnotesize
\begin{tabular}{@{}l r@{}}
\toprule
Similarity threshold & Fraction of test notes \\
\midrule
0.30 & 6.20\% \\
0.50 & 1.50\% \\
0.70 & 0.60\% \\
0.85 & 0.20\% \\
\\
\bottomrule
\end{tabular}
\end{table}

\begin{table}[!htbp]
\centering
\caption{Leakage audit checklist (TeMLM-Datasheet).}
\label{tab:leak_check}
\footnotesize
\begin{tabularx}{\linewidth}{@{}lX@{}}
\toprule
Check & Description \\
\midrule
Patient overlap & Confirm unique patient identifiers are disjoint across splits \\
Near-duplicate text & Similarity scan on note text; report thresholds and rates \\
Label leakage & Verify features/metadata do not encode ground truth (e.g., templated headers) \\
Contamination & Ensure evaluation sets are not used in instruction tuning/prompt selection \\
\bottomrule
\end{tabularx}
\end{table}

\subsection{Mapping to clinical AI reporting guidelines}
TeMLM is not a replacement for guideline-driven reporting, but a structured way to generate the evidence those guidelines require. Table~\ref{tab:mapping} maps TeMLM artifacts to common reporting expectations.

\begin{table*}[!t]
\centering
\caption{Mapping TeMLM artifacts to major clinical AI reporting guidance.}
\label{tab:mapping}
\footnotesize
\begin{tabularx}{0.98\textwidth}{@{}lXXX@{}}
\toprule
Guideline & Transparency pain point & TeMLM artifact(s) & Example output \\
\midrule
CONSORT-AI / SPIRIT-AI \cite{liu2020consortai,liu2020spiritai} & Data provenance and AI component description in trials & Datasheet + Provenance & Versioned data pipeline; reproducible evaluation runs \\
TRIPOD+AI \cite{collins2024tripodai} & Complete reporting of prediction model development \& validation & Model card + Provenance & Subgroup performance; calibration; drift monitoring plan \\
DECIDE-AI \cite{vasey2022decideai} & Early-stage clinical evaluation and human factors & Model card & Intended workflow; oversight plan; failure modes \& mitigation \\
CLAIM \cite{mongan2020claim} & Reproducible model and dataset reporting & Datasheet + Model card & Standardized disclosures; audit-ready artifacts \\
\bottomrule
\end{tabularx}
\end{table*}

\section{Discussion}
Transparency mechanisms are sometimes treated as supplemental material appended after model development, but in clinical AI they function as part of the safety case: they determine whether results are interpretable, reproducible, and fit for deployment. The TeMLM transparency pillar reframes ``documentation'' as infrastructure by (i) making disclosures versioned artifacts, (ii) binding those artifacts to concrete data/model events via provenance, and (iii) attaching measurable audit targets (completeness, contamination risk, privacy residual risk, and reliability).

\subsection{Transparency as reviewer-grade evidence}
Peer review in clinical NLP often fails at the same point: reviewers cannot verify whether splits were patient-level, whether label generation was stable, whether de-identification was validated beyond tool outputs, or whether evaluation was inadvertently contaminated. TeMLM addresses this by turning narrative claims into machine-checkable fields that can be inspected and diffed across versions. Practically, this reduces the reporting burden for authors: once datasheet and provenance records exist, many reporting items become auto-fillable, and the remaining narrative can focus on clinical motivation and empirical results.

\subsection{Alignment with clinical AI reporting and governance}
Reporting guidelines such as CONSORT-AI, SPIRIT-AI, TRIPOD+AI, and DECIDE-AI specify what to report, but they do not prescribe a portable encoding that enables verification or reuse \cite{liu2020consortai,liu2020spiritai,collins2024tripodai,vasey2022decideai}. TeMLM complements these checklists by providing a structured representation for the same evidence: dataset lineage, preprocessing assumptions, label provenance, subgroup evaluations, and monitoring plans. This matters for governance: drift monitoring, incident response, and update policies cannot be credibly assessed if the underlying pipeline is opaque \cite{challen2019safety,char2018implementing,kelly2019key}. TeMLM-Provenance links monitoring outputs (e.g., PSI alerts) to the exact dataset version and transformation chain that produced them, enabling root-cause analysis rather than post-hoc speculation.

\subsection{Cumulative science under privacy constraints}
Clinical text is difficult to share; many groups cannot redistribute raw data even when their models and claims matter clinically. Transparency artifacts help close this gap by allowing others to audit assumptions and reproduce transformations on local data. In effect, TeMLM shifts the unit of reproducibility from ``the dataset'' to ``the pipeline and its disclosures.'' This supports multi-site replication and negative results: performance changes can be traced to differences in annotation policy, documentation templates, or patient mix rather than attributed to ``randomness.''

\subsection{Avoiding transparency theater}
A legitimate concern is ``transparency theater'': producing rich documentation that is not verified, not updated, or disconnected from the actual training run. TeMLM reduces this risk in two ways. First, provenance binds disclosures to concrete events (extraction, de-identification, splitting, training, evaluation) so that outdated narratives are easier to detect. Second, TeMLM treats key transparency properties as measurable gates. For example, documentation completeness is tracked quantitatively (Fig.~\ref{fig:datasheet_completeness}); leakage audits are reported as full curves rather than single thresholds (Fig.~\ref{fig:leak_curve}); and annotation reliability is required alongside human labels (Fig.~\ref{fig:deid_benchmark}). The point is not to claim that any single metric guarantees safety, but to ensure that omissions and high-risk conditions are visible and actionable.

\subsection{Relationship to TeMLM explainability and evaluation pillars}
Transparency is necessary but not sufficient for trustworthy medical language modeling. Explainability methods (attribution, evidence retrieval, rationale generation) can be persuasive even when they are unfaithful or poorly evaluated \cite{bender2021dangers,raji2020closinggap}. In the broader TeMLM program, transparency provides the substrate for (i) explanation faithfulness testing (linking explanations to evidence and provenance), (ii) governance and safety processes (risk taxonomies and incident handling), and (iii) clinical evaluation protocols that measure workflow impact rather than offline accuracy alone.

\subsection{Limitations and future directions}
This preprint has several limitations, which also define a concrete research agenda.
\begin{itemize}
  \item \textbf{Synthetic dataset scope.} The worked example uses \emph{Technetium-I}, a fully synthetic corpus designed for reproducible auditing and benchmarking. While this supports privacy-preserving development and process validation, synthetic notes may not capture institution-specific jargon, rare identifiers, or complex longitudinal narratives. Deployment-facing claims should therefore be validated on governed real-world clinical data.
  \item \textbf{Metric minimalism vs. clinical adequacy.} The proposed metrics are intentionally minimal. They do not yet cover calibration under distribution shift for generative outputs, causal validity of proxy labels, or clinically meaningful error taxonomies for summarization and question answering. Extending the metric suite should be guided by intended use and risk, consistent with clinical reporting frameworks \cite{collins2024tripodai,vasey2022decideai}.
  \item \textbf{Engineering cost and incentives.} Provenance capture is easiest when pipelines are designed for it; retrofitting provenance into legacy systems can be costly. Future work should quantify the implementation burden, identify automation opportunities (e.g., automatic extraction of datasheet fields from ETL logs), and evaluate whether structured artifacts improve review outcomes and downstream reuse.
  \item \textbf{Transparency trade-offs.} Some transparency goals conflict: aggressive de-identification can reduce utility, and strict access control can reduce external scrutiny. TeMLM does not resolve these trade-offs, but it requires them to be stated explicitly (threat model, verification protocol, and release constraints) so that reviewers and users can evaluate whether the risk posture matches the claimed application.
\end{itemize}

Overall, TeMLM argues that ``transparent enough'' should be treated as a verifiable claim: documentation, provenance, and audit metrics together form a portable evidence bundle for clinical NLP.

\section{Conclusion}
We presented the Transparency pillar of TeMLM, introducing TeMLM-Datasheet, TeMLM-Card, and TeMLM-Provenance together with a minimal transparency metric suite and release thresholds. These artifacts are intended to raise the floor for medical language model reporting and to make clinical NLP research more auditable, reusable, and trustworthy.

\section*{Author contributions (CRediT)}
Conceptualization: OYLI, TY, ATT, OG. Methodology: OYLI, TY, ATT, OG, MNZ. Software: OYLI, MNZ, DUKU, IO. Validation: DE, SBD, RIT, OAK. Investigation: all authors. Writing--original draft: OYLI, TY, ATT, OG. Writing--review \& editing: all authors. Supervision: OYLI, TY, ATT, OG.

\section*{Data and code availability}
The worked example uses the \emph{Technetium-I} synthetic clinical NLP dataset \cite{technetium_i_2026}. The TeMLM templates and provenance schema are intended for open release; this preprint includes all figures generated from the dataset's audit statistics. Technetium-I is distributed under the European Union Public Licence v1.2 (EUPL-1.2) \cite{eupl12}.

\section*{Declaration of competing interest}
The authors declare no competing interests.

\bibliographystyle{elsarticle-num}
\bibliography{references}

\end{document}